\documentclass[sigconf]{acmart}
\usepackage{multirow}
\usepackage{algorithm}
\usepackage{algorithmic}
\usepackage{booktabs}
\usepackage{float}
\usepackage{amsmath}
\usepackage{enumerate}
\usepackage{graphicx} 
\usepackage{array}
\usepackage{bm}   
\usepackage{stfloats}
\renewcommand\footnotetextcopyrightpermission[1]{} 
\usepackage{balance} 

\theoremstyle{remark}
\newtheorem{remark}{Remark}
\newtheorem{theorem}{Theorem}

\newtheorem{lemma}{Lemma}
\newtheorem{corollary}{Corollary}

\DeclareMathAlphabet\mathbfcal{OMS}{cmsy}{b}{n}

\AtBeginDocument{%
  \providecommand\BibTeX{{%
    \normalfont B\kern-0.5em{\scshape i\kern-0.25em b}\kern-0.8em\TeX}}}

\copyrightyear{2025}
\acmYear{2025}
\setcopyright{acmlicensed}\acmConference[MM '25]{Proceedings of the 33rd ACM International Conference on Multimedia}{October 27--31, 2025}{Dublin, Ireland}
\acmBooktitle{Proceedings of the 33rd ACM International Conference on Multimedia (MM '25), October 27--31, 2025, Dublin, Ireland}
\acmDOI{10.1145/3746027.3755396}
\acmISBN{979-8-4007-2035-2/2025/10}

\begin{document}

\title{LargeMvC-Net: Anchor-based Deep Unfolding Network for Large-scale Multi-view Clustering} 

\author{Shide Du}
\email{dushidems@gmail.com}
\affiliation{%
  \institution{Fuzhou University}
  \city{Fuzhou}
  \country{China}
}

\author{Chunming Wu}
\email{chunmingwu0102@163.com}
\affiliation{%
  \institution{Fuzhou University}
  \city{Fuzhou}
  \country{China}
}

\author{Zihan Fang}
\email{fzihan11@163.com}
\affiliation{%
  \institution{Fuzhou University}
  \city{Fuzhou}
  \country{China}
}

\author{Wendi Zhao}
\email{241010030@fzu.edu.cn}
\affiliation{%
  \institution{Fuzhou University}
  \city{Fuzhou}
  \country{China}
}

\author{Yilin Wu}
\email{a767220005@gmail.com}
\affiliation{%
  \institution{Fuzhou University}
  \city{Fuzhou}
  \country{China}
}

\author{Changwei Wang}
\email{changweiwang@sdas.org}
\affiliation{%
  \institution{Shandong Academy of Sciences}
  \city{Jinan}
  \country{China}
}

\author{Shiping Wang}
\email{shipingwangphd@163.com}
\affiliation{%
  \institution{Fuzhou University}
  \city{Fuzhou}
  \country{China}
}
\authornote{Corresponding author.}

\renewcommand{\shortauthors}{Shide Du et al.}

\begin{abstract}
Deep anchor-based multi-view clustering methods enhance the scalability of neural networks by utilizing representative anchors to reduce the computational complexity of large-scale clustering.
Despite their scalability advantages, existing approaches often incorporate anchor structures in a heuristic or task-agnostic manner, either through post-hoc graph construction or as auxiliary components for message passing.
Such designs overlook the core structural demands of anchor-based clustering, neglecting key optimization principles.
To bridge this gap, we revisit the underlying optimization problem of large-scale anchor-based multi-view clustering and unfold its iterative solution into a novel deep network architecture, termed LargeMvC-Net.
The proposed model decomposes the anchor-based clustering process into three modules: RepresentModule, NoiseModule, and AnchorModule, corresponding to representation learning, noise suppression, and anchor indicator estimation.
Each module is derived by unfolding a step of the original optimization procedure into a dedicated network component, providing structural clarity and optimization traceability.
In addition, an unsupervised reconstruction loss aligns each view with the anchor-induced latent space, encouraging consistent clustering structures across views.
Extensive experiments on several large-scale multi-view benchmarks show that LargeMvC-Net consistently outperforms state-of-the-art methods in terms of both effectiveness and scalability.
The source data and code are available.\footnote{\scriptsize\url{https://github.com/dushide/LargeMvC-Net\_ACMMM\_2025}}
\end{abstract}



\maketitle

\section{Introduction}\label{sec:int}

Unsupervised learning on multi-view data has become a cornerstone problem in modern machine learning, with applications spanning cross-modal retrieval \cite{Lin25Testtime}, sensor fusion \cite{Lupion243Dhuman}, and multi-modal recommendation \cite{Lin23Contrastive}.
In this context, multi-view clustering seeks to discover intrinsic cluster structures by leveraging consistency and complementary information across views \cite{Luo24Fairgt, Kou24Exploiting, Luo24Fugnn, Kou25Instance}.
A vital challenge in multi-view clustering lies in the need to jointly model diverse feature distributions while maintaining global structural consistency, all without supervision.
As datasets grow in size and modality complexity, method scalability become increasingly crucial for practical deployment.

\begin{figure}[t]
  \centering
  \includegraphics[width=\linewidth]{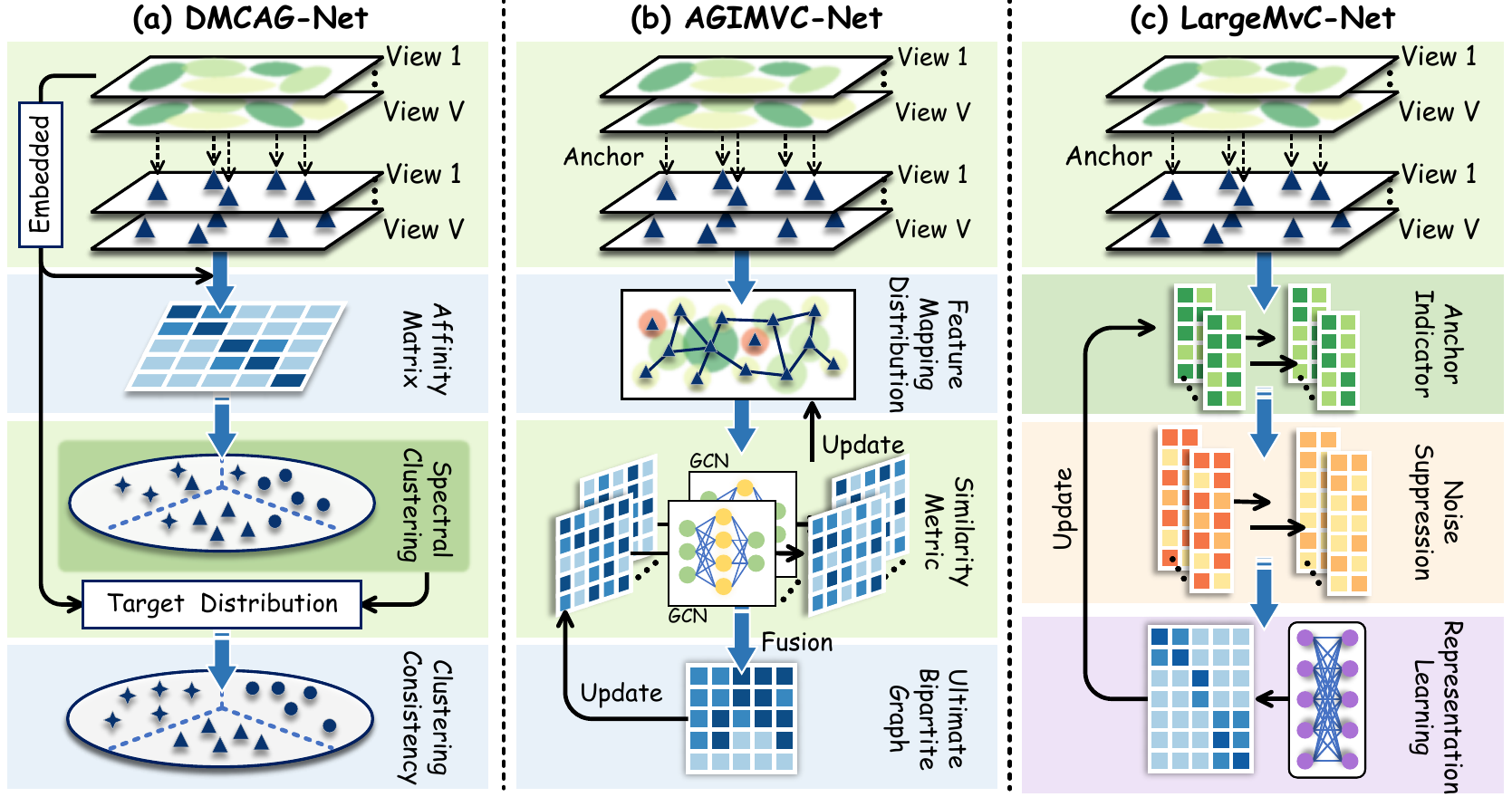}\\
  \caption{Existing SOTA deep anchor-based clustering methods (DMCAG-Net \cite{Cui23Deepmultiview} and AGIMVC-Net \cite{Fu23Anchorgraph}) \textit{vs.} The proposed LargeMvC-Net.
  Here, they often lack interpretability, and their optimization is detached from the underlying clustering principle.}
  \label{Framework0}
\end{figure}

Recent efforts toward scalable multi-view clustering have turned to shallow anchor-based methods \cite{Liu22Efficient, Wang2024Scalable, Li24Fastunpaired, Wang25Multimodal, Wang25Bidirectional}, which approximate sample-level relationships using a compact set of representative anchors.
These methods construct anchor graphs or indicator matrices to capture local structure and reduce computational overhead, achieving impressive scalability on large-scale benchmarks.
However, such models often rely on shallow linear formulations, which limits their ability to encode deep semantic correlations or handle modality-specific corruptions effectively.
Their resulting clustering representations are frequently too coarse to capture rich cross-view alignment.
To solve this problem, deep multi-view clustering methods \cite{Wang21Generative, Cui24Dualcontrast, Xu24Deepvariational, Kou24Inaccurate, Wang25HighlyEfficient} are proposed to adopt neural architectures to learn expressive latent embeddings through joint feature transformation and alignment.
They offer improved modeling flexibility and are capable of capturing complex non-linear relationships.
Nevertheless, these approaches typically require access to full data and global similarity computation, making them computationally prohibitive for large-scale scenarios.
Fortunately, recent work has been attempted on combining deep networks with anchor-based shallow methods, such as DMCAG-Net \cite{Cui23Deepmultiview} and AGIMVC-Net \cite{Fu23Anchorgraph}.
Despite the advancements in deep anchor-based multi-view clustering, which incorporate anchor structures to approximate local relationships and reduce computational complexity, significant challenges remain.
Existing methods often incorporate anchor structures in a heuristic or task-agnostic manner, either through post-hoc graph construction (DMCAG-Net \cite{Cui23Deepmultiview}) or as auxiliary components for message passing (AGIMVC-Net \cite{Fu23Anchorgraph}), as shown in Fig. \ref{Framework0}.
This leads to a failure in directly integrating the core structural demands of anchor-based clustering into the network's learning process. 
As a result, it prevents the preservation of important optimization principles from the original formulation.

\begin{figure*}[t]
  \centering
  \includegraphics[width=\linewidth]{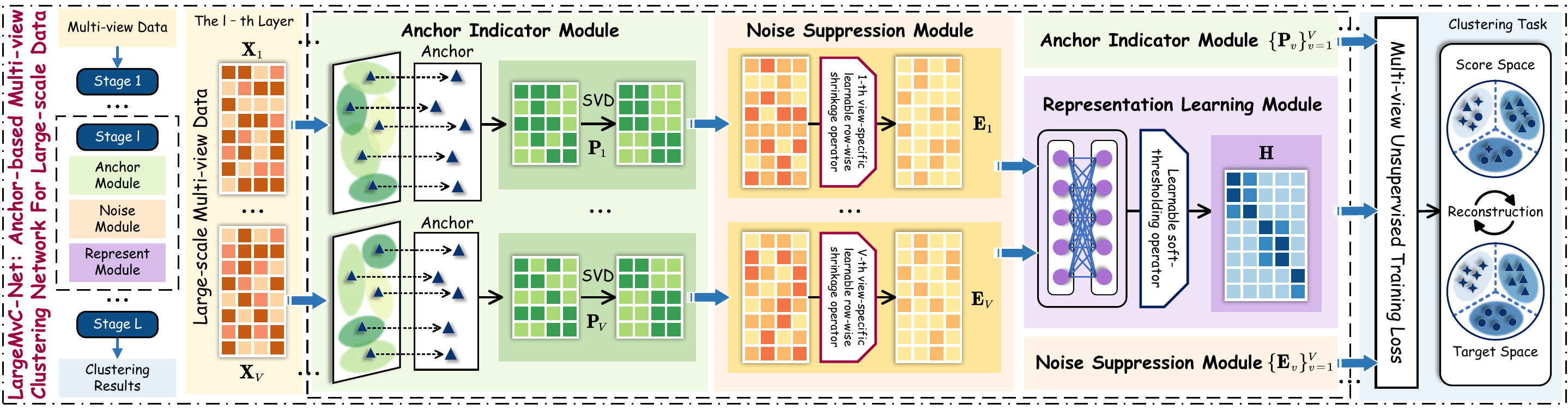}\\
  \caption{An overview of the proposed anchor-based deep unfolding network (LargeMvC-Net).}
  \label{Framework}
\end{figure*}

To bridge this gap, we propose a novel framework, \textbf{LargeMvC-Net}, that integrates the scalability of anchor-based models with the expressive power of deep architectures.
Our approach revisits the optimization problem of large-scale anchor-based multi-view clustering and unfolds its iterative solution into a structured deep network.
Rather than relying on generic representation networks such as autoencoders or GCNs and applying anchor structures indirectly through post-hoc graph construction or message passing, as in \cite{Cui23Deepmultiview, Fu23Anchorgraph}, we directly translate each step of the original optimization. 
This includes clustering representation update, noise suppression, and anchor alignment, into a dedicated network module.
The resulting architecture comprises three interpretable components: \textbf{RepresentModule} for multi-view consistent representation learning, \textbf{NoiseModule} for adaptive view-specific denoising, and \textbf{AnchorModule} for orthogonally-constrained anchor alignment.
This design preserves the structural insights of the original formulation while enabling end-to-end training via unsupervised reconstruction loss.
The overall framework is outlined in Fig. \ref{Framework}, and the main contributions of this paper can be listed as follows:
\begin{itemize}
\item \textbf{\textit{Formulation of LargeMvC-Net:}} We propose LargeMvC-Net, a deep unfolding network tailored for large-scale multi-view clustering with anchor guidance.
\item \textbf{\textit{Anchor-based optimization-inspired network design:}} We derive a principled architecture by unfolding the optimization steps of a robust anchor-based clustering formulation into modular and interpretable network components.
\item \textbf{\textit{Extensive experiments on large-scale benchmarks:}} Extensive experiments on large-scale multi-view datasets show that LargeMvC-Net outperforms state-of-the-art shallow and deep methods in both clustering quality and scalability.
\end{itemize}

\section{Related Work}\label{RelatedWork}

\textbf{Anchor-based Multi-view Clustering.} 
1) \textbf{Complete shallow anchor-based multi-view clustering} constructs anchor indicator matrices from a small set of instances.
This aims to approximate large affinity graphs with improved efficiency.
For example, Chen \textit{et al.} \cite{Chen22Efficient} jointly optimized anchor learning, graph construction, and large-scale clustering for better multi-view representation.
Chen \textit{et al.} \cite{Chen24Concept} integrated anchor learning, semantic coefficient representation, and partitioning while explicitly modeling multi-view data.
Ji \textit{et al.} \cite{Ji2025Anchors} introduced large-scale anchor representation learning into non-convex low-rank tensor learning with enhanced tensor rank, consistent geometric regularization, and tensorial exclusive regularization.
2) \textbf{Incomplete shallow anchor-based multi-view clustering} builds compact anchor matrices under missing data settings, achieving efficient storage with preserved clustering performance.
For instance, Wang \textit{et al.} \cite{Wang2022highly} integrated multi-view anchor learning and incomplete bipartite graphs to perform large-scale clustering and introduced a flexible bipartite graph approach.
Li \textit{et al.} \cite{Li23Distribution} leveraged a consensus latent space, anchor-based fast imputation, and distribution consistency, while enforcing a tensor low-rank constraint for high-order correlation exploration.
Du \textit{et al.} \cite{Du24Fastand} refined the bipartite graph structure by optimizing an anchor-side graph filter within a large-scale incomplete consensus clustering framework.
Further work on shallow anchor-based multi-view clustering can be discovered in \cite{Ji23Anchorstructure, Li24TensorizedLabel, Gu2024Dictionary, Wang25Pseudo, Wang25Incompletetensorized} (complete) and \cite{Wen23ScalableIncomplete, Li23Crossview, He23Structured, Liu24Alleviate, Li24ParameterFree} (incomplete).
3) \textbf{Deep anchor-based multi-view clustering} combines anchor-based structural approximation with deep networks to achieve scalable and expressive clustering across views. 
Recent methods can be seen in DMCAG-Net \cite{Cui23Deepmultiview} and AGIMVC-Net \cite{Fu23Anchorgraph}.

\textbf{Deep Multi-view Clustering.}
1) \textbf{Complete deep multi-view clustering} uses deep networks to learn shared representations, capturing consistency and complementarity across views for improved clustering.
For example, Wang \textit{et al.} \cite{Wang2023Triple} introduced a contrastive learning framework that leveraged multi-view autoencoders and affinity fusion to enhance deep subspace clustering performance.
Zhao \textit{et al.} \cite{Zhao24DFMVC}  ensured fair multi-view clustering by learning deep consistent representations and aligning sensitive attributes with cluster distribution.
Wang \textit{et al.} \cite{Wang24MultiViewsubspace} presented a structured multi-pathway network for deep multi-view clustering, integrating multilevel features through a shared connection matrix with a low-rank constraint.
2) \textbf{Incomplete deep multi-view clustering} ensures robustness to missing views via imputation-free learning or adaptive feature alignment.
For instance, Yang \textit{et al.} \cite{Yang2022Robust} introduced a noise-robust contrastive loss to mitigate false negatives from random sampling, offering a unified solution for incomplete deep multi-view clustering.
Xu \textit{et al.} \cite{Xu23Adaptive} proposed an imputation-free deep incomplete multi-view clustering method that learns view-specific features via autoencoders and aligned feature distributions through adaptive projection.
Pu \textit{et al.} \cite{Pu24Adaptive} addressed incomplete multi-view clustering by employing deep encoders for feature extraction, constructing latent graphs to preserve structural information.
More work could also be seen in \cite{Chen23Joint, Wang23Adversarial, Wang24SURER, Du2025Openviewer} (complete) and \cite{Yang21Partially, Tang22Deep, Jin23Deepalignment, Lin23Dualcontrastive, Du2024UMCGL, Kou25Progressive} (incomplete).

\textbf{Deep Unfolding Network.}
Deep unfolding networks offer interpretable architectures by structurally mirroring the step-by-step logic of the underlying optimization problems.
They have achieved success in multiple fields \cite{GregorL10, Marivani20Multimodal, Bonet22Explaining, Zheng23HybridISTA, Boris24InterpretableNN}.
For example, Luong \textit{et al.} \cite{Luong21Designing} presented a deep recurrent neural networks, designed by the unfolding of iterative algorithms that solved the task of sequential video reconstruction.
Du \textit{et al.} \cite{Du23Bridging} bridged trustworthiness with deep unfolding networks for open-set learning to enhance design-level interpretability.             
Weerdt \textit{et al.} \cite{Weerdt24DeepUnfolding} designed an optimization problem for sequential signal recovery and derived into a deep unfolding transformer network architecture.
More similar attempts could also be traced in \cite{Arrieta2020Explainable, Ning21Accurate, Yang22Memoryaugmented}.


\section{The Proposed Framework}\label{ProposedFramework}

\begin{table}[!htbp]
\centering
\caption{Essential notations and descriptions.}
\resizebox{\linewidth}{!}{
\begin{tabular}{c||c}
\toprule {Notations} & {Descriptions} \\ 
\midrule
   {$\mathbb{R}$}   & The real number space.\\
   {$V$, $n$}   & The number of views and samples.\\
   {$d_{v}$}   & The dimension of the $v$-th view feature.\\
   {$c$, $m$}   & The number of clusters and anchors.\\
  \midrule
   {$\{\mathbf{X}_{v}\}_{v=1}^{V}$}   & $\mathbf{X}_{v} \in \mathbb{R}^{n \times d_{v}}$ is the $v$-view sample matrix. \\
   {$\{\mathbf{E}_{v}\}_{v=1}^{V}$}   & $\mathbf{E}_{v} \in \mathbb{R}^{n \times d_{v}}$ is the $v$-th noise suppression matrix.\\
    {$\{\mathbf{P}_{v}\}_{v=1}^{V}$}   & $\mathbf{P}_{v} \in \mathbb{R}^{m \times d_{v}}$ is the $v$-th anchor indicator matrix.\\
   {$\mathbf{H}$}   & $\mathbf{H} \in \mathbb{R}^{n \times m}$ is the consistent clustering representation.\\
  \bottomrule
\end{tabular}}
\label{SymbolicNormalization}
\end{table}

\subsection{Multi-view Anchor-based Problem Formulation and Optimization}\label{MVAProblem}
The necessary notations are first listed in Table \ref{SymbolicNormalization}.
To enable scalable multi-view clustering on large-scale data, we adopt an anchor-based formulation that approximates global sample-level similarity using a small set of representative anchors. 
This approach circumvents the high computational cost of full pairwise similarity computation while preserving critical structural information across views.

\textbf{Optimization Foundation for Deep Unfolding}: Let $\{\mathbf{X}_{v} \in \mathbb{R}^{n \times d_{v}}\}_{v=1}^{V}$ denote the input multi-view data, where $n$ is the number of samples and $d_{v}$ is the feature dimensionality of the $v$-th view.
Our goal is to obtain a large-scale consistent clustering representation $\mathbf{H}$ over $m \ll n$ anchors, such that information from all $V$ views is aligned in a unified low-dimensional latent space. 
To this end, we introduce a set of anchors that serve as representative structural proxies for the original data distribution.
Each view-specific reconstruction is mediated by an anchor indicator matrix $\mathbf{P}_{v} \in \mathbb{R}^{m \times d_{v}}$.
This maps anchor-based representations to the original view space.
In this consideration, the anchor mechanism not only reduces computational cost but also establishes a shared structural basis across heterogeneous views. 
This allows for cross-view alignment through anchor indicators. 
Based on this, a fundamental optimization problem arises in the anchor-based multi-view clustering task. 
The goal is to minimize the aggregated reconstruction error across all views while ensuring sparsity in the shared representation as
\begin{equation}\label{allproblem0}
\begin{array}{ll}
\mathop{\min}\limits_{\mathbf{H}, \mathbf{P}_{v}}\sum\limits_{{v=1}}^{V}\Big(\frac{1}{2}\Vert\mathbf{X}_{v}-\mathbf{H}\mathbf{P}_{v}\Vert_{F}^{2}+\alpha \Vert\mathbf{H}\Vert_{1}\Big), \text { s.t. } \mathbf{H} \geq 0, \mathbf{P}_{v}\mathbf{P}_{v}^{\top}=\mathbf{I},
\end{array}
\end{equation}
where the scalar $\alpha > 0$ balances the trade-off between sparsity and reconstruction fidelity.
Herein, the first term reconstructs the input $\mathbf{X}_{v}$ using the consistent representation $\mathbf{H}$ and anchor indicator $\mathbf{P}_{v}$.
The $\ell_1$-norm regularization enforces sparsity, enhancing its discriminative capacity for clustering. 
The non-negativity constraint $\mathbf{H} \geq 0$ ensures that each sample’s representation is composed only of positive contributions from latent anchors, preserving the additive nature of cluster assignment.
While the orthogonality constraint $\mathbf{P}_{v}\mathbf{P}_{v}^{\top}=\mathbf{I}$ stabilizes the anchor indicator and avoids trivial solutions.

However, since the basic formulation in Problem~\eqref{allproblem0} assumes that each view can be reconstructed solely through shared representations and orthogonal projections.
It becomes vulnerable to view-specific corruptions commonly observed in real-world multi-view data, such as sensor failures, occlusions, or modality-dependent noise.
To overcome this limitation, we introduce a view-specific noise term $\mathbf{E}_{v}$ for each view and adopt an additional $\ell_{2, 1}$-norm regularization to separate structured noise. 
The resulting problem is reconstructed as
\begin{equation}\label{allproblem}
\begin{array}{ll}
\mathop{\min}\limits_{\mathbf{H}, \mathbf{P}_{v}, \mathbf{E}_{v}}&\sum\limits_{{v=1}}^{V}\Big(\frac{1}{2}\Vert\mathbf{X}_{v}-\mathbf{H}\mathbf{P}_{v}-\mathbf{E}_{v}\Vert_{F}^{2}+\alpha \Vert\mathbf{H}\Vert_{1} + \beta \Vert\mathbf{E}_{v}\Vert_{2, 1}\Big), \\& \text { s.t. } \mathbf{H} \geq 0, \mathbf{P}_{v}\mathbf{P}_{v}^{\top}=\mathbf{I},
\end{array}
\end{equation}
where the scalar $\beta > 0$ balances the trade-off between noise removal and reconstruction fidelity.
Problem~\eqref{allproblem} jointly learns a consistent clustering representation $\mathbf{H}$, view-specific anchor indicator matrices $\mathbf{P}_{v}$, and structured noise estimations $\mathbf{E}_{v}$, to enable scalable and corruption-tolerant cross-view alignment.
By solving the objective in Problem~\eqref{allproblem}, the model aims to learn a consistent clustering representation $\mathbf{H}$ across multiple views.
It leverages view-specific anchor indicator $\mathbf{P}_{v}$ to align heterogeneous feature spaces and incorporates structured noise terms $\mathbf{E}_{v}$ to filter out view-specific sample-level corruptions.
To solve Problem~\eqref{allproblem}, we adopt an alternating minimization strategy that iteratively updates $\mathbf{H}$, $\mathbf{P}_{v}$, and $\mathbf{E}_{v}$ while keeping the others fixed. 
Each sub-problem admits efficient closed-form or proximal updates, as described below.

\textbf{1) Optimization with Respect to $\mathbf{H}$}: Given fixed anchor indicator matrices $\mathbf{P}_{v}$ and noise matrices $\mathbf{E}_{v}$, we update the consistent clustering representation $\mathbf{H}$, which encodes the latent assignments of samples to anchors. 
This sub-problem is described as
\begin{equation}\label{Zproblem1}
\begin{array}{ll}
\mathop{\min}\limits_{\mathbf{H}}\sum\limits_{{v=1}}^{V}\Big(\frac{1}{2}\Vert\mathbf{X}_{v}-\mathbf{H}\mathbf{P}_{v}-\mathbf{E}_{v}\Vert_{F}^{2}+\alpha \Vert\mathbf{H}\Vert_{1}\Big), \text { s.t. } \mathbf{H} \geq 0.
\end{array}
\end{equation}
This is a $\ell_1$-norm regularized least squares problem with a non-negativity constraint. 
It can be efficiently solved using a proximal gradient method with soft-thresholding operator $\mathbfcal{S}_{\lambda_1}(\cdot)$, where $\lambda_1$ is a $\ell_{1}$-norm sparsity threshold.
The iterative update is given by
\begin{equation}\label{Zsolveproblem1}
\begin{array}{ll}
\mathbf{H}^{(l+1)} \leftarrow \frac{1}{V}\sum\limits_{{v=1}}^{V}&\Bigg(\mathbfcal{S}_{\frac{\alpha}{L_{p}}}\Big(\mathbf{H}^{(l)}-\frac{1}{L_{p}}(\mathbf{H}^{(l)}\mathbf{P}_{v}^{(l)}(\mathbf{P}_{v}^\top)^{(l)}\\&-\mathbf{X}_{v}(\mathbf{P}_{v}^\top)^{(l)}+\mathbf{E}_{v}^{(l)}(\mathbf{P}_{v}^\top)^{(l)})\Big)\Bigg),
\end{array}
\end{equation}
where $l$ is the current iteration number, and $L_{p}$ denotes the Lipschitz constant of the gradient with respect to $\mathbf{H}$.
Alternatively, this can be re-expressed in a more structured form as
\begin{equation}\label{Zsolveproblem2}
\begin{array}{ll}
\mathbf{H}^{(l+1)} \leftarrow &\frac{1}{V}\sum\limits_{{v=1}}^{V}\Big(\mathbfcal{S}_{\frac{\alpha}{L_{p}}}\Big(\mathbf{H}^{(l)}(\mathbf{I}-\frac{1}{L_{p}}\mathbf{P}_{v}^{(l)}(\mathbf{P}_{v}^\top)^{(l)})\\&+\frac{1}{L_{p}}(\mathbf{X}_{v}-\mathbf{E}_{v}^{(l)})(\mathbf{P}_{v}^\top)^{(l)}\Big).
\end{array}
\end{equation}

\textbf{2) Optimization with Respect to $\mathbf{E}_{v}$}: With fixing $\mathbf{H}$ and $\mathbf{P}_{v}$, the update of the view-specific noise matrix $\mathbf{E}_{v}$ reduces to solving the following sub-problem as
\begin{equation}\label{Evproblem1}
\begin{array}{ll}
\mathop{\min}\limits_{\mathbf{E}_{v}}\frac{1}{2}\Vert\mathbf{X}_{v}-\mathbf{H}\mathbf{P}_{v}-\mathbf{E}_{v}\Vert_{F}^{2}+ \beta \Vert\mathbf{E}_{v}\Vert_{2, 1}.
\end{array}
\end{equation}
This is a standard row-sparse minimization problem, where the $\ell_{2, 1}$-norm encourages the removal of view-specific sample-level corruptions. 
The optimal solution can be obtained via the row-wise shrinkage operator $\mathbfcal{D}_{\lambda_2}(\cdot)$, where $\lambda_2$ is a $\ell_{2, 1}$-norm sparsity threshold.
Accordingly, the related update becomes
\begin{equation}\label{Esolveproblem2}
\begin{array}{ll}
\mathbf{E}_{v}^{(l+1)} \leftarrow \mathbfcal{D}_{\frac{\beta}{L_{q_{v}}}}\left(\mathbf{X}_{v} - \mathbf{H}^{(l+1)}\mathbf{P}_{v}^{(l)}\right),
\end{array}
\end{equation}
where $L_{q_{v}}$ denotes the $v$-th Lipschitz constant of the gradient with respect to $\mathbf{E}_{v}$.

\textbf{3) Optimization with Respect to $\mathbf{P}_{v}$}: Finally, with $\mathbf{H}$ and $\mathbf{E}_{v}$ fixed, the anchor indicator matrix $\mathbf{P}_{v}$ is updated by solving the following orthogonally-constrained least squares sub-problem as
\begin{equation}\label{Dvproblem1}
\begin{array}{ll}
\mathop{\min}\limits_{\mathbf{P}_{v}}\frac{1}{2}\Vert\mathbf{X}_{v}-\mathbf{H}\mathbf{P}_{v}-\mathbf{E}_{v}\Vert_{F}^{2}, \text { s.t. } \mathbf{P}_{v}\mathbf{P}_{v}^{\top}=\mathbf{I}.
\end{array}
\end{equation}
Inspired by \cite{Chen24Concept, Wang2024Scalable}, this sub-problem can be equivalently rewritten as the following trace maximization problem as
\begin{equation}\label{Dsolveproblem1}
\begin{array}{ll}
\mathop{\max}\limits_{\mathbf{P}_{v}}\operatorname{Tr}\left(\mathbf{P}_{v}^\top(\mathbf{H}^{\top}\mathbf{X}_{v}-\mathbf{H}^{\top}\mathbf{E}_{v})\right), \text { s.t. } \mathbf{P}_{v}\mathbf{P}_{v}^{\top}=\mathbf{I}.
\end{array}
\end{equation}
Problem~\eqref{Dsolveproblem1} admits a closed-form solution based on the orthogonal Procrustes problem \cite{Wang19Multi}.  
Specifically, we perform the singular value decomposition (SVD) as 
\begin{equation}\label{Dsolveproblem2}
\begin{array}{ll}
\operatorname{\textbf{SVD}}((\mathbf{H}^{\top})^{(l+1)}\mathbf{X}_{v}-(\mathbf{H}^{\top})^{(l+1)}\mathbf{E}_{v}^{(l+1)})=\mathbf{B}_{v}^{(l+1)}\mathbf{\Sigma}_{v}(\mathbf{C}_{v}^{\top})^{(l+1)},
\end{array}
\end{equation}
where $\mathbf{B}_{v} \in \mathbb{R}^{m \times m}$, $\mathbf{\Sigma}_{v}\in \mathbb{R}^{m \times m}$ and $\mathbf{C}_{v}\in \mathbb{R}^{d_v \times m}$.
By using the left and right singular value matrices $\mathbf{B}_{v}$ and $\mathbf{C}_{v}$, the next iteration of $\mathbf{P}_{v}$ can be updated.

\subsection{Anchor-based Deep Unfolding Clustering Network Architecture}
While the alternating optimization framework in Subsection~\ref{MVAProblem} provides an interpretable and scalable solution to anchor-based multi-view clustering, but lacks the expressive capacity of deep models.
To integrate the strengths of classical anchor-based optimization process and deep representation learning, we propose a principled deep unfolding architecture inspired by \cite{Arrieta2020Explainable, Ning21Accurate, Yang22Memoryaugmented}, unfolding the optimization steps into a feed-forward network structure.
This results in a layer-wise architecture composed of three interpretable components: \textbf{RepresentModule}, \textbf{NoiseModule}, and \textbf{AnchorModule}.
The overall architecture unfolds for $L$ stages (layers), with each stage mimicking one iteration of the anchor-based optimization learning routine.

\textbf{1) Representation Learning Module (RepresentModule)}: This module is derived from the update rule of $\mathbf{H}$ in Eq.~\eqref{Zsolveproblem2}, and is responsible for updating the clustering representation. 
It aggregates multi-view residuals and suppresses irrelevant components via a learnable soft-thresholding operation as
\begin{equation}\label{Zmodule}
\begin{array}{ll}
\mathbf{H}^{(l+1)} \leftarrow \frac{1}{V}\sum\limits_{{v=1}}^{V}\left(\mathbfcal{S}_{\theta^{(l)}}\Big(\mathbf{H}^{(l)}\mathbf{R}+(\mathbf{X}_{v}-\mathbf{E}_{v}^{(l)})(\mathbf{P}_{v}^\top)^{(l)}\mathbf{U}\right),
\end{array}
\end{equation}
where $\mathbf{R}\in \mathbb{R}^{m \times m} = \mathbf{I}-\frac{1}{L_{p}}\mathbf{P}_{v}\mathbf{P}_{v}^\top$ and $\mathbf{U} \in \mathbb{R}^{m \times m} = \frac{1}{L_{p}}\mathbf{I}$ are two trainable network layers, and $\mathbf{I}$ is an identity matrix. 
$\mathbfcal{S}_{\theta}(\boldsymbol{a}^{(ij)})=\sigma(\boldsymbol{a}^{(ij)}-\theta)-\sigma(-\boldsymbol{a}^{(ij)}-\theta)$ denotes a soft-thresholding operator with learnable threshold $\theta = \frac{\alpha}{L_{p}}$, enabling flexible sparsity control.
$\boldsymbol{a}^{(ij)}$ is the element in the $i$-th row and $j$-th column of the matrix, $\boldsymbol{a}^{(i)}$ is the $i$-th column of the matrix, and $\sigma(\cdot)$ can be activation functions such as ReLU, SeLU and etc.
\textbf{RepresentModule} \eqref{Zmodule} captures cross-view latent clustering alignment and promotes representation learning.

\textbf{2) Noise Suppression Module (NoiseModule)}: This module implements the update of view-specific corruption matrices $\mathbf{E}_{v}$, following the form of Eq.~\eqref{Esolveproblem2}. 
It aims to isolate sample-level noise from meaningful reconstruction signals as
\begin{equation}\label{Evmodule}
\begin{array}{ll}
\mathbf{E}_{v}^{(l+1)}\leftarrow\mathbfcal{D}_{\rho_{v}^{(l)}}\left(\mathbf{X}_{v}- \mathbf{H}^{(l+1)}\mathbf{P}_{v}^{(l)}\right),
\end{array}
\end{equation}
where $\mathbfcal{D}_{\rho_{v}}(\boldsymbol{a}^{(i)}) = \frac{\sigma(\left\|\boldsymbol{a}^{(i)}\right\|_2-\rho_{v})}{\left\|\boldsymbol{a}^{(i)}\right\|_2} \boldsymbol{a}^{(i)}$, 
if $\rho_{v}<\left\|\boldsymbol{a}^{(i)}\right\|_2$; otherwise, $0$, which is the $v$-th view-specific learnable row-wise shrinkage operator with threshold parameter $\rho_{v} = \frac{\beta}{L_{q_{v}}}$, generalized from the $\ell_{2, 1}$-norm proximal operator.
\textbf{NoiseModule} \eqref{Evmodule} adaptively suppresses noisy or corrupted samples in each view and improves the robustness of LargeMvC-Net.

\textbf{3) Anchor Indicator Module (AnchorModule)}: 
AnchorModule is responsible for estimating the view-specific anchor indicator matrix $\mathbf{P}_{v}$ based on SVD-based update in Eq.~\eqref{Dsolveproblem2}. 
This step ensures that each view’s anchor structural pattern is properly aligned with the shared latent space as
\begin{equation}\label{Dvmodule}
\begin{array}{ll}
\mathbf{P}_{v}^{(l+1)}=\mathbf{B}_{v}^{(l+1)}(\mathbf{C}_{v}^{\top})^{(l+1)},
\end{array}
\end{equation}
where $\mathbf{B}_{v}$ and $\mathbf{C}_{v}$ are the left and right singular matrix.
\textbf{AnchorModule} \eqref{Dvmodule} aligns each view’s anchor structure to the current clustering representation, ensuring structure-preserving latent modeling under orthogonality constraints.

\begin{algorithm}[t]
\caption{LargeMvC-Net}
\label{algorithmLargeMvCNet}
\begin{algorithmic}[1]
\REQUIRE{Multi-view data $\{\mathbf{X}_{v}\}_{v=1}^{V}$, training epochs $T$, the number of unfolding networks $L$, the amount of anchors $m$, and learning rate $\eta$}.
\ENSURE {$\mathbf{H}$ as the anchor-based representation for $k$-means.}
\STATE {Initialize network parameters $\mathbf{\Theta}=\{\mathbf{R}, \mathbf{U}, \theta, \rho_{v}\}$;}
\STATE {Initialize the $v$-th anchor matrix $\mathbf{P}_{v}$ by $k$-means;}
\FOR {$t = 1 \rightarrow T$}
\FOR {$l = 1 \rightarrow L$}
\STATE {Obtain the anchor-based representation $\mathbf{H}^{(l)}$ by RepresentModule \eqref{Zmodule};}
\STATE {Calculate the noise matrix $\mathbf{E}_{v}^{(l)}$ by NoiseModule \eqref{Evmodule};}
\STATE {Update anchor matrix $\mathbf{P}_{v}^{(l)}$ by AnchorModule \eqref{Dvmodule};}
\ENDFOR
\STATE {Compute the multi-view unsupervised training loss \eqref{lossRe};}
\STATE {Update $\mathbf{\Theta}$ though backward propagation;}
\ENDFOR
\RETURN{$\mathbf{H}$ as the anchor-based representation for $k$-means.}
\end{algorithmic}
\end{algorithm}

\subsection{Multi-view Unsupervised Training Loss} 
To enable unsupervised training of LargeMvC-Net, we introduce a reconstruction-based training loss that leverages the anchor-induced clustering representation as the structural association. 
The critical idea is to ensure that the clustering representation $\mathbf{H}$, learned through deep unfolding.
It is sufficiently expressive to reconstruct each view’s original input through the learned view-specific anchor indicator matrices $\mathbf{P}_{v}$.
Specifically, given the $t$-th training epoch reconstruction $\hat{\mathbf{X}}_{v}^{(t)} = \mathbf{H}^{(t)}\mathbf{P}_{v}^{(t)}$, we minimize the reconstruction loss between each view’s input and its corresponding reconstruction as
\begin{equation}
\begin{array}{ll}\label{lossRe}
\mathcal{L}_{R} = \sum\limits_{{v=1}}^{V}\textbf{MSE}(\mathbf{X}_{v}, \hat{\mathbf{X}}_{v}^{(t)}) =\sum\limits_{{v=1}}^{V}(\|\mathbf{X}_{v}-\mathbf{H}^{(t)}\mathbf{P}_{v}^{(t)}\|_{F}^{2}).
\end{array}
\end{equation}
Training loss \eqref{lossRe} encourages the anchor-based latent space $\mathbf{H}$ to preserve sufficient information for reconstructing all views.
Meanwhile, it ensures that view-specific anchor indicators $\mathbf{P}_{v}$ are aligned with the clustering representation. 
Unlike contrastive or pseudo-label based approaches, our reconstruction loss avoids reliance on handcrafted construction tasks or noisy clustering signals.
Additional discussions on loss scalability can be seen in \textbf{Appendix} Subsection \ref{DiscussiononTrainingLosses}.
We summarize the complete end-to-end training procedure of LargeMvC-Net in Algorithm~\ref{algorithmLargeMvCNet}.


\subsection{Theoretical Analysis}

\begin{table}[t]
\centering
\caption{Complexity on SOTA shallow methods, where $d$ is an intermediate variable dimension, $d \ll n$.}
\resizebox{\linewidth}{!}{
\begin{tabular}{c||c|c}
\toprule {Methods} & {Time Complexity} & {Space Complexity} \\ 
\midrule
   {LMVSC (AAAI'20) \cite{Kang2020a}}   & $\mathcal{O}(n(mV+mD+c^2)+m^3V^3)$ & $\mathcal{O}(Vc(n+D))$ \\
   {AIMC (ACMMM'22) \cite{Chen22Adaptively}}   & $\mathcal{O}(nmc + d^2D + dc^2D + dc^3)$ & $\mathcal{O}(nc + d(c + D))$\\
   {FMVACC (NIPS'22) \cite{Wang22Alignthen}}   & $\mathcal{O}(n(mD + m^2) + m^2D + m^3)$ & $\mathcal{O}(mD + nmV + m^2V)$ \\
   {AWMVC (AAAI'23) \cite{Wan23AutoWeighted}}   & $\mathcal{O}(n(c^2 + d^2 + dD) + dc^2)$ & $\mathcal{O}(n(c^2 + d^2) + dc^2)$ \\
   {EMVGC-LG (ACMMM'23) \cite{Wen23EfficientMulti}}   & $\mathcal{O}(n(mDV+m^3)+m^3V)$ & $\mathcal{O}(nm+mD)$ \\
   {FastMICE (TKDE'23) \cite{Huang23FastMulti}}   & $\mathcal{O}(ncm^{\frac{1}{2}}V^{\frac{1}{2}})$ & $\mathcal{O}(n(c+m+K+V))$\\
   {FDAGF (AAAI'23) \cite{Zhang23Let}}   & $\mathcal{O}(n(mD+m^2)+m^2D)$ & $\mathcal{O}(n(D+m)+mD)$ \\
   {MVSC-HFD (IF'24) \cite{Ou24Anchorbased}}   & $\mathcal{O}(n(cD+mc)+mcD+cdD)$ & $\mathcal{O}(n(c+m+d)+mD+cD)$ \\
   {RCAGL (TKDE'24) \cite{Liu24Robust}}   & $\mathcal{O}(n(m^2+md))$ & $\mathcal{O}(n(dV+mV))$\\
   {UDBGL (TNNLS'24) \cite{Fang24EfficientMulti}}   & $\mathcal{O}(n(mD+mc+m^2D+mV^2)+m^3)$ & $\mathcal{O}(n(m+D)+mD)$ \\
  \midrule
  {IMVC-CBG (CVPR'22) \cite{Wang2022highly}}   & $\mathcal{O}(n(cd+mc+md)+mcd)$ & $\mathcal{O}(n(d+m)+mc+cd)$\\
  {SIMVS-SA (ACMMM'23) \cite{Wen23ScalableIncomplete}}   & $\mathcal{O}(n(md+m^2V)+m^3V+m^2d)$ & $\mathcal{O}(n(d+m)+md)$\\
  {FIMVC (TNNLS'24) \cite{Liu24FastIncomplete}}   & $\mathcal{O}(n(m^2+md)+m^2d)$ & $\mathcal{O}(n(d+m)+md)$\\
  {FSIMVC-OF (ACMMM'24) \cite{Du24Fastand}}   & $\mathcal{O}(n(mc+m^2+md+mc^2)+m^3)$ & $\mathcal{O}(n(c+m+d))$\\
  {PSIMVC-PG (TNNLS'24) \cite{Li24ParameterFree}}   & $\mathcal{O}(n(cd+c^2)+c^2d)$ & $\mathcal{O}(n(d+c)+c^2)$\\
  \midrule
  {LargeMvC-Net (Ours)}   & $\mathcal{O}(L(n(mD + m^2V) + m^2D))$ & $\mathcal{O}(n(m + D) + mD + m^2V)$ \\
  \bottomrule
\end{tabular}}
\label{ShallowComplexity}
\end{table}

\subsubsection{Convergence Analysis}\label{ConvergenceAnalysis}
Here, we provide a brief convergence proof of LargeMvC-Net based on the following \textbf{Theorem \ref{Theorem1}}. 
\begin{theorem}\label{Theorem1} 
Given that the objective function $\mathcal{J}(\cdot)$ is lower bounded by zero and monotonically non-increasing (by \textbf{Lemma \ref{Lemma1})}, the proposed LargeMvC-Net is guaranteed to converge.
\end{theorem}
Furthermore, although some variables are parameterized as learnable components, they are optimized using gradient descent as training progresses. 
This ensures the following \textbf{Corollary \ref{Corollary1}}.
\begin{corollary}\label{Corollary1}
The convergence resulting in \textbf{Theorem \ref{Theorem1}} still holds in the presence of learnable parameters, provided that they are updated using appropriate gradient-based optimization techniques.
\end{corollary}
\begin{remark}\label{Remark1}
The above analysis provides important theoretical support for deploying LargeMvC-Net in real-world multi-view scenarios. 
In practice, the data often comes from heterogeneous sources with noise, redundancy, or view-specific corruption. 
The guaranteed monotonic decrease of the objective function ensures that the model's training process remains stable, even when some views are incomplete or contain low-quality information. 
Moreover, the convergence to a minimum ensures that the network will not oscillate or diverge during training. 
This property is particularly crucial when applying the unfolding model to large-scale multi-view clustering.
Thus, these theoretical properties enhance the reliability and robustness of LargeMvC-Net in practical applications \cite{Bezdek03Convergence}.
Additional proofs can be found in \textbf{Appendix} Section \ref{SupplementaryofOverallConvergence}.

\end{remark}

\subsubsection{Computational Complexity Analysis}\label{ComputationalComplexityAnalysis}

\begin{table}[t]
\centering
\caption{Complexity on SOTA deep methods with $L$ layers, where $b$ is the mini-batch size, and $s$ is the maximum number of neurons in the hidden layers. ``Max reported" means the largest dataset size reported in the original paper.}
\resizebox{\linewidth}{!}{
\begin{tabular}{c||c|c}
\toprule {Methods} & {Time Complexity} & {Max Reported} \\ 
\midrule
   {SDSNE-Net (AAAI'22) \cite{Liu22Stationary}}   & $\mathcal{O}(Ln^3)$ & 18,758 \\
   {CVCL-Net (ICCV'23) \cite{Chen23Contrasting}}   & $\mathcal{O}(L(nbsD+n^2b^2c^2+nbcD))$ & 10,000 \\
   {SCMVC-Net (TMM'24) \cite{Wu24Selfweighted}}   & $\mathcal{O}(L(n^2sV+nsc))$ & 50,000 \\
   \midrule
   {DMCAG-Net (IJCAI'23) \cite{Cui23Deepmultiview}}   & $\mathcal{O}(L(n(sD+msV+m^2+cV^2)))$ & 10,000 \\
   \midrule
   {DIMVC-Net (AAAI'22) \cite{Xu22Deep}}   & $\mathcal{O}(L(n(cD+sV)+c^3))$ & 4,485 \\
   {IRDMC-Net (TNNLS'24) \cite{Liu23Information}}   & $\mathcal{O}(L(n^2D+nsV^2+nsc))$ & 10,800 \\
    \midrule
    {AGIMVC-Net (TNNLS'23) \cite{Fu23Anchorgraph}}   & $\mathcal{O}(L(n(mD+sD+sm)))$ & 126,054 \\
  \midrule
  {LargeMvC-Net (Ours)}   & $\mathcal{O}(L(n(mD + m^2V) + m^2D))$ & 195,537 \\
  \bottomrule
\end{tabular}}
\label{DeepComplexity}
\end{table}

The overall computational complexity is dominated by RepresentModule, which integrates information across $V$ views. 
Specifically, it requires $\mathcal{O}(nmD + nm^2V)$ time and $\mathcal{O}(nm + nD + mD + m^2)$ space, where $n$ is the number of samples, $m$ is the number of anchors, and $D = \sum_{v=1}^V d_v$ denotes the total feature dimensions across views. 
NoiseModule performs feature-wise residual denoising with a cost of $\mathcal{O}(nmD)$ time and $ \mathcal{O}(nD + mD)$ space. 
For AnchorModule, we adopt an SVD-based mechanism by reconstructing $\mathbf{P}_v$ via the left and right singular vectors of a view-specific matrix. 
This yields a complexity of $ \mathcal{O}(m^2D)$ in time and $\mathcal{O}(m^2V + mD)$ in space.  
Summing across all components, the total time complexity is $\mathcal{O}(nmD + nm^2V+ m^2D)$, and the total space complexity is $\mathcal{O}(nm + nD + mD + m^2V)$.
When we consider iterating $L$ deep unfolding layers, its complexity in the $t$-th training epoch extends to: the total time complexity costs $\mathcal{O}(L(n(mD + m^2V) + m^2D))$.
In practice, since $m$, $V$, and $L$ are typically much smaller than $n$, that is $m, V, L \ll n$, in real-world multi-view applications, the overall complexity scales linearly with the number of samples, \emph{i.e.,} $\mathcal{O}(n)$.
As a result, the proposed method maintains linear complexity, ensuring practical scalability for real-world large-scale multi-view applications.

\begin{remark}\label{Remark2}
Here, we present a theoretical comparison of the major computational complexities in Tables~\ref{ShallowComplexity}-\ref{DeepComplexity} to provide a fair assessment of algorithmic efficiency.
In fact, in large-scale scenarios, the time complexity of these methods can be approximately regarded as $\mathcal{O}(n)$, but the space complexity is slightly different.

\textbf{Differences from Previous Shallow and Deep Anchor-based Multi-view Methods.} Traditional anchor-based multi-view clustering methods improve scalability by replacing dense graphs with compact anchor structures \cite{Chen22Efficient, Ji23Anchorstructure, Chen24Concept, Li24TensorizedLabel, Ji2025Anchors}. 
However, they rely on shallow linear assumptions that fail to capture nonlinear cross-view dependencies \cite{Gu2024Dictionary, Li23Distribution, Wang2022highly}.
Recent deep models such as DMCAG-Net \cite{Cui23Deepmultiview} and AGIMVC-Net \cite{Fu23Anchorgraph} incorporate anchors into neural networks to enhance robustness and scalability. 
Nevertheless, they lack principled integration with clustering objectives and optimization logic.
While auxiliary losses (\emph{e.g.,} entropy regularization or neighborhood-preserving constraints) can be incorporated, the core anchor-based reconstruction objective alone yields strong and stable performance across benchmarks (see Subsection~\ref{sec:exprq}).
\end{remark}

\section{Experiments and Analyses}\label{sec:experiments}

\subsection{Experimental Setups}\label{ExperimentalSetup}

\subsubsection{Datasets, Compared Methods, and Evaluation Metrics} 

\begin{table}[t]
\centering
\caption{A brief description of the tested datasets.}
\resizebox{0.95\linewidth}{!}{
\begin{tabular}{c||c|c|c|c}
\toprule
Datasets & \# Samples & \# Views & \# Feature Dimensions & \# Classes \\
\midrule
Animals & 10,158 & 2 & 4,096/4,096 & 50 \\
Caltech102  & 9,144  & 6 & 48/40/254/1,984/512/928 & 102  \\
Cifar10  & 50,000  & 3 & 2,048/512/1024 & 10 \\
MNIST & 60,000 & 3 & 342/64/1024 & 10 \\
NUSWIDEOBJ  & 30,000 & 5 & 65/226/145/74/129 & 31 \\
YoutubeFace & 101,499 & 5 & 64/512/64/647/838 & 31 \\
YTF-50 & 126,054 & 4 & 944/576/512/640 & 50 \\
YTF-100 & 195,537 & 4 & 944/576/512/640 & 50 \\
\midrule
ESP-Game & 11,032 & 2 & 100/100 & 7 \\
Flickr & 12,154 & 2 & 100/100 & 7 \\
IAPR & 7,855 & 2 & 100/100 & 6 \\
\bottomrule
\end{tabular}}
\label{Datadescription}
\end{table}

We conduct experiments in challenging large-scale multi-view clustering tasks under well-known multi-view datasets, including: Animals, Caltech102, Cifar10, MNIST, NUSWIDEOBJ, YoutubeFace, YTF-50, YTF-100, ESP-Game, Flickr, and IAPR.
These datasets have two types, feature-level and modality-level scenarios: 1) Animals, Caltech102, Cifar10, MNIST, NUSWIDEOBJ, YoutubeFace, YTF-50 and YTF-100 datasets contain different manual and deep features; 2) ESP-Game, Flickr, and IAPR datasets include various vision and language features.
Moreover, Animals, Caltech102, NUSWIDEOBJ, YoutubeFace, ESP-Game, and IAPR are also performed in the incomplete multi-view setting.
Here, we randomly apply a sample missing rate $\{0.1, 0.3, \cdots, 0.9\}$ to each view, while ensuring every sample retains at least one complete view.
The statistics of these datasets are summarized in Table \ref{Datadescription} (details in \textbf{Appendix} Subsection \ref{SupDatasetsandTestSettings}).

To verify the superiority of LargeMvC-Net, several large-scale complete and incomplete multi-view clustering models are introduced. 
Complete shallow methods involve: LMVSC \cite{Kang2020a}, AIMC \cite{Chen22Adaptively}, FMVACC \cite{Wang22Alignthen}, AWMVC \cite{Wan23AutoWeighted}, EMVGC-LG \cite{Wen23EfficientMulti}, FastMICE \cite{Huang23FastMulti}, FDAGF \cite{Zhang23Let}, MVSC-HFD \cite{Ou24Anchorbased}, RCAGL \cite{Liu24Robust}, and UDBGL \cite{Fang24EfficientMulti}.
Complete deep methods have: SDSNE-Net \cite{Liu22Stationary}, CVCL-Net \cite{Chen23Contrasting}, SCMVC-Net \cite{Wu24Selfweighted}, and anchor-based DMCAG-Net \cite{Cui23Deepmultiview}.
Meanwhile, incomplete shallow methods contain: IMVC-CBG \cite{Wang2022highly}, SIMVS-SA \cite{Wen23ScalableIncomplete}, FIMVC \cite{Liu24FastIncomplete}, FSIMVC-OF \cite{Du24Fastand}, and PSIMVC-PG \cite{Li24ParameterFree}.
Incomplete deep methods include: DIMVC-Net \cite{Xu22Deep}, IRDMC-Net \cite{Liu23Information}, and anchor-based AGIMVC-Net \cite{Fu23Anchorgraph}.
Herein, we rely on source code provided by the authors, and tune to the best performance as in their papers for fair comparison.
Moreover, we utilize three commonly used metrics to evaluate clustering performance: Clustering accuracy (ACC), normalized mutual information (NMI), and adjusted rand index (ARI).

\subsubsection{Implementation Details}\label{ImplementationDetails}
LargeMvC-Net is implemented using the PyTorch on an NVIDIA GeForce RTX 3090 GPU with 24GB memory. 
Moreover, we train the network for 100 epochs with a learning rate of 0.01.
Meanwhile, we search $L$ in $[1, 2, \cdots, 9]$ and $m$ in $[c, 2c, 3c]$, $m$ is the anchor number and $c$ is the cluster number.
Each experiment is conducted ten times, and the means and standard deviation are reported as the final results.
The ablation-models (\textbf{Appendix} Table \ref{NetworkExamples}) are RMvC-Net (w/o AnchorModule and NoiseModule) and AMvC-Net (w/o NoiseModule).

\subsection{Experimental Results}\label{sec:exprq}

\subsubsection{Complete and Incomplete Clustering Results}
\begin{table*}[t]
\centering
\caption{Clustering results of large-scale compared multi-view clustering methods, where the best and runner-up performance are highlighted in \textbf{\color{red}red} and \textbf{\color{blue}blue} respectively (mean\% and standard deviation\%). "–" indicates the out-of-memory error.}
\resizebox{0.98\textwidth}{!}{
\begin{tabular}{c|c||c|cc|cccc|ccc||ccc|c|c}
\toprule
\multicolumn{2}{c||}{Datasets $\backslash$ Methods} & LMVSC & AIMC & FMVACC & AWMVC & EMVGC-LG & FastMICE & FDAGF & MVSC-HFD & RCAGL & UDBGL & SDSNE-Net & CVCL-Net & SCMVC-Net & DMCAG-Net & LargeMvC-Net\\
\multicolumn{2}{c||}{(Publisher'Year)}    &  (AAAI’20) & (MM’22) & (NIPS’22) & (AAAI’23) & (MM’23) & (TKDE’23) & (AAAI’23) & (IF’24) & (TKDE’24) & (TNNLS’24) & (AAAI’22) & (ICCV’23) & (TMM’24)  & (IJCAI’23) & (Ours) \\
\midrule
\multirow{3}*{Animals} & ACC       & 27.91 (0.00) & 58.09 (0.00) & 53.53 (0.79) & 65.36 (0.00) & 62.82 (1.33) & 64.39 (3.67) & 59.69 (0.00) & 52.29 (0.00) & \textbf{\color{blue}65.69 (0.00)} & 65.30 (0.00) & 56.08 (4.38) & 30.56 (0.03) & 31.69 (0.00) & 59.90 (0.00) & \textbf{\color{red}68.01 (0.77)} \\
                    & NMI       & 35.35 (0.00) & 68.75 (0.00) & 62.88 (1.03) & \textbf{\color{blue}71.73 (0.00)} & 70.91 (1.27) & 66.48 (2.63) & 69.93 (0.00) & 65.38 (0.00) & 71.71 (0.00) & 70.49 (0.00) & 68.45 (2.48) & 44.10 (0.02) & 46.17 (0.00) & 65.80 (0.00) & \textbf{\color{red}71.80 (0.32)} \\
                    & ARI       & 15.43 (0.00) & 49.66 (0.00) & 44.23 (0.59) & 46.12 (0.00) & \textbf{\color{blue}52.69 (2.17)} & 47.84 (2.16) & 49.50 (0.00) & 44.47 (0.00) & 46.17 (0.00) & 48.12 (0.00) & 23.87 (6.25) & 18.71 (0.02) & 22.98 (0.00) & 42.71 (0.00) & \textbf{\color{red}53.44 (1.07)} \\
\midrule
\multirow{3}*{Caltech102} & ACC    & 11.66 (0.00) & 23.94 (0.00) & 20.54 (1.30) & 29.14 (0.00) & 27.15 (0.26) & 20.05 (1.73) & 30.03 (0.00) & 20.24 (0.00) & 36.67 (0.00) & 19.95 (0.00) & \textbf{\color{blue}42.69 (0.65)} & 17.43 (0.00) & 21.40 (0.00) & 22.57 (0.00) & \textbf{\color{red}48.64 (0.00)} \\
                    & NMI       & 25.31 (0.00) & 34.35 (0.00) & 40.76 (0.92) & 50.88 (0.00) & \textbf{\color{blue}51.03 (1.88)} & 44.48 (1.58) & 51.02 (0.00) & 31.50 (0.00) & 48.44 (0.00) & 33.37 (0.00) & 48.09 (0.79) & 38.03 (0.00) & 42.10 (0.00) & 27.39 (0.00) & \textbf{\color{red}51.48 (0.00)} \\
                    & ARI       & 2.38 (0.00) & 13.57 (0.00) & 15.06 (0.62) & 25.50 (0.00) & 20.31 (2.90) & 15.52 (2.44) & 24.32 (0.00) & 10.96 (0.00) & 21.72 (0.00) & \textbf{\color{blue}43.20 (0.00)} & 14.68 (3.07) & 13.70 (0.00) & 20.80 (0.00) & 6.89 (0.00) & \textbf{\color{red}51.91 (0.00)} \\
\midrule
\multirow{3}*{Cifar10}  & ACC  & 88.98 (0.00) & 98.32 (0.00) & 88.69 (2.73) & 96.90 (0.00) & 96.28 (1.07) & 98.11 (2.45) & 96.27 (0.00) & 98.09 (0.00) & \textbf{\color{blue}98.98 (0.00)} & 64.57 (0.00) & - & 98.24 (0.01) & 98.16 (0.00) & - & \textbf{\color{red}99.13 (0.00)} \\
                    & NMI       & 79.17 (0.00) & 97.12 (0.00) & 81.36 (2.01) & 92.68 (0.00) & 91.78 (0.66) & 97.15 (1.90) & 91.32 (0.00) & 97.01 (0.00) & \textbf{\color{blue}97.29 (0.00)} & 81.57 (0.00) & - & 95.58 (0.01) & 97.12 (0.00) & - & \textbf{\color{red}97.64 (0.00)} \\
                    & ARI       & 76.93 (0.00) & 97.50 (0.00) & 78.61 (1.46) & 93.39 (0.00) & 92.39 (0.41) & 97.04 (1.22) & 92.04 (0.00) & 97.00 (0.00) & \textbf{\color{blue}97.58 (0.00)} & 51.85 (0.00) & - & 96.17 (0.01) & 97.17 (0.00) & - & \textbf{\color{red}98.10 (0.00)} \\
\midrule
\multirow{3}*{NUSWIDEOBJ}& ACC   & 10.69 (0.00) & 19.33 (0.00) & 11.88 (1.18) & 12.88 (0.00) & 12.57 (0.85) & 14.80 (1.41) & 13.37 (0.00) & 18.01 (0.00) & 19.22 (0.00) & 13.32 (0.00) & \textbf{\color{blue}24.73 (0.75)} & 15.63 (0.00) & 18.97 (0.00) & 13.15 (0.00) & \textbf{\color{red}27.79 (0.21)} \\
                    & NMI       & 8.28 (0.00) & 13.29 (0.00) & 10.26 (2.29) & 12.01 (0.00) & 11.70 (1.70) & 14.24 (0.46) & 12.46 (0.00) & 12.31 (0.00) & 12.42 (0.00) & 10.76 (0.00) & 12.55 (1.07) & 13.45 (0.00) & 13.75 (0.00) & \textbf{\color{blue}14.43 (0.00)} & \textbf{\color{red}14.72 (0.18)} \\
                    & ARI       & 1.72 (0.00) & 6.66 (0.00) & 3.17 (0.27) & 4.02 (0.00) & 3.60 (1.95) & 5.29 (0.83) & 4.06 (0.00) & 6.02 (0.00) & 5.63 (0.00) & 3.73 (0.00) & 7.36 (1.35) & 7.35 (0.00) & \textbf{\color{blue}9.31 (0.00)} & 2.74 (0.00) & \textbf{\color{red}10.83 (0.35)} \\
\midrule
\multirow{3}*{YTF-50} & ACC  & 60.57 (0.00) & 68.02 (0.00) & 69.50 (3.18) & 73.34 (0.00) & 69.85 (2.38) & 70.39 (1.65) & 67.86 (0.00) & 67.81 (0.00) & \textbf{\color{blue}75.28 (0.00)} & 64.11 (0.00) & - & 64.81 (0.00) & 73.72 (0.00) & - & \textbf{\color{red}80.13 (0.65)} \\
                    & NMI    & 78.66 (0.00) & 84.70 (0.00) & 83.01 (2.33) & \textbf{\color{blue}85.45 (0.00)} & 84.17 (2.14) & 83.32 (2.07) & 82.21 (0.00) & 82.76 (0.00) & 84.49 (0.00) & 80.73 (0.00) & - & 79.11 (0.00) & 82.76 (0.00) & - & \textbf{\color{red}85.53 (0.25)} \\
                    & ARI       & 49.44 (0.00) & 65.52 (0.00) & 59.35 (1.90) & \textbf{\color{blue}66.94 (0.00)} & 62.26 (1.16) & 61.25 (0.98) & 56.56 (0.00) & 61.30 (0.00) & 63.66 (0.00) & 53.18 (0.00) & - & 57.04 (0.00) & 65.31 (0.00) & - & \textbf{\color{red}67.68 (1.10)} \\
\midrule
\multirow{3}*{YTF-100} & ACC    & 53.83 (0.00) & 66.84 (0.00) & 65.36 (1.06) & \textbf{\color{blue}67.90 (0.00)} & 64.99 (0.00) & 63.91 (0.93) & 61.83 (1.99) & 62.92 (0.00) & 67.49 (0.00) & 61.50 (0.00) & - & 58.84 (0.01) & 67.66 (0.00) & - & \textbf{\color{red}76.52 (0.20)} \\
                   & NMI        & 76.83 (0.00) & 83.28 (0.00) & 82.72 (2.42) & \textbf{\color{blue}84.41 (0.00)} & 83.45 (1.35) & 82.95 (0.76) & 80.80 (0.00) & 81.18 (0.00) & 83.07 (0.00) & 79.67 (0.00) & - & 79.19 (0.00) & 81.13 (0.00) & - & \textbf{\color{red}84.59 (0.09)} \\
                   & ARI        & 34.69 (0.00) & 55.31 (0.00) & 53.77 (2.61) & 58.08 (0.00) & 55.89 (2.31) & 58.04 (0.80) & 46.09 (0.00) & 52.60 (0.00) & 58.09 (0.00) & 32.84 (0.00) & - & 50.72 (0.01) & \textbf{\color{blue}61.39 (0.00)} & - & \textbf{\color{red}62.46 (1.16)} \\
\midrule
\midrule
\multirow{3}*{Flickr}  & ACC  & 41.93 (0.00) & 49.59 (0.00) & 52.51 (1.73) & 51.93 (0.00) & 52.51 (1.01) & 51.60 (2.93) & 45.02 (0.00) & 50.67 (0.00) & 50.06 (0.00) & 40.97 (0.00) & 48.44 (3.06) & \textbf{\color{blue}52.68 (0.03)} & 52.21 (0.00) & 34.70 (0.00) & \textbf{\color{red}53.37 (0.06)} \\
                    & NMI        & 23.11 (0.00) & 32.40 (0.00) & 32.81 (1.97) & 33.63 (0.00) & 33.91 (1.77) & 31.77 (2.60) & 29.35 (0.00) & 27.80 (0.00) & 34.26 (0.00) & 33.85 (0.00) & 34.35 (2.19) & \textbf{\color{blue}34.42 (0.03)} & 30.97 (0.00) & 20.98 (0.00) & \textbf{\color{red}34.72 (0.05)} \\
                    & ARI        & 13.98 (0.00) & 27.57 (0.00) & 27.54 (1.83) & 28.78 (0.00) & 28.57 (1.19) & 27.27 (1.21) & 19.98 (0.00) & 24.08 (0.00) & 29.25 (0.00) & 17.32 (0.00) & 22.66 (3.31) & \textbf{\color{blue}29.27 (0.05)} & 28.54 (0.00) & 10.89 (0.00) & \textbf{\color{red}29.72 (0.04)} \\
\midrule
\multirow{3}*{IAPR}   & ACC & 38.97 (0.00) & 38.50 (0.00) & 38.32 (2.84) & 44.16 (0.00) & 43.89 (2.16) & 45.33 (2.07) & 33.97 (0.00) & 35.28 (0.00) & 44.43 (0.00) & 38.87 (0.00) & 36.27 (2.86) & \textbf{\color{blue}46.61 (0.03)} & 44.12 (0.00) & 34.69 (0.00) & \textbf{\color{red}46.97 (0.53)} \\
                    & NMI        & 16.90 (0.00) & 19.19 (0.00) & 17.96 (2.97) & 23.39 (0.00) & 22.58 (2.10) & 24.30 (1.86) & 17.43 (0.00) & 16.84 (0.00) & 23.74 (0.00) & 22.39 (0.00) & 19.43 (3.67) & \textbf{\color{blue}24.54 (0.03)} & 22.67 (0.00) & 18.67 (0.00) & \textbf{\color{red}24.95 (0.72)} \\
                    & ARI        & 13.04 (0.00) & 14.66 (0.00) & 13.73 (3.03) & 18.41 (0.00) & 17.91 (1.93) & 18.04 (1.53) & 11.67 (0.00) & 11.27 (0.00) & 18.81 (0.00) & 15.76 (0.00) & 11.18 (2.35) & \textbf{\color{blue}19.24 (0.03)} & 18.58 (0.00) & 11.82 (0.00) & \textbf{\color{red}19.68 (0.26)} \\
\bottomrule
\end{tabular}}
\label{AllcomparsionClustering}
\end{table*}

\begin{figure*}[t]
  \centering
  \includegraphics[width=0.94\linewidth]{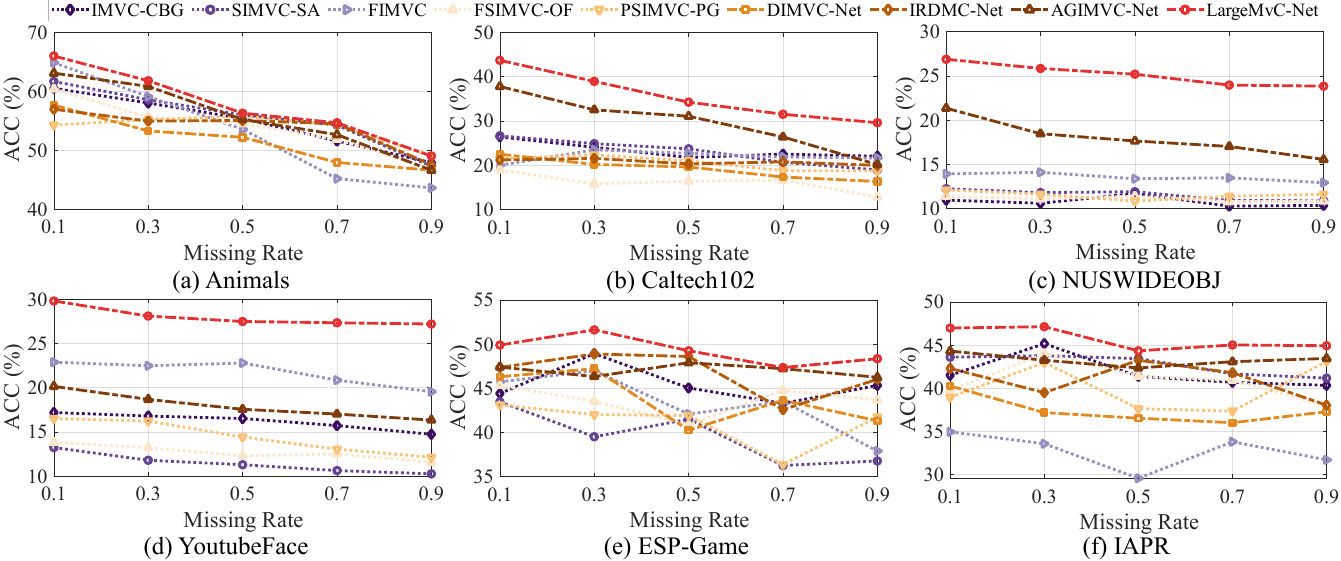}\\
  \caption{ACC of large-scale incomplete clustering methods on multi-view datasets with different missing rates.}
  \label{AcccomparsionInClustering}
\end{figure*}

Table~\ref{AllcomparsionClustering} and Fig.~\ref{AcccomparsionInClustering} present results on large-scale multi-view benchmarks, where LargeMvC-Net consistently achieves top performance across under complete and incomplete conditions.
\textbf{a) Shallow \textit{vs.} Ours.} Shallow methods like RCAGL, UDBGL, IMVC-CBG, and SIMVC-SA often rely on handcrafted anchors or static bipartite graphs, which perform poorly in the presence of noisy inputs or missing views. 
In contrast, LargeMvC-Net integrates anchor optimization into a scalable deep unfolding framework, enabling robust alignment and semantic preservation across both complete and incomplete settings.
\textbf{b) Deep \textit{vs.} Ours.} While deep baselines such as CVCL-Net and DIMVC-Net offer strong representation learning, they lack explicit structural modeling. 
Even anchor-based deep models like DMCAG-Net and AGIMVC-Net incorporate anchor graphs in a heuristic or task-agnostic manner, decoupled from the clustering objective. 
LargeMvC-Net unfolds the full anchor-based optimization process into interpretable modules, achieving superior structure-aware clustering at scale.
\textbf{c) Dataset Trends.} For feature-level multi-view datasets (\emph{e.g.,} YTF-50, YTF-100), our model excels under both complete and incomplete views, showing stability to noise and partial features. 
On modality-level multi-view datasets (\emph{e.g.,} Flickr, IAPR, ESP-Game), LargeMvC-Net generalizes well without modality-specific designs, reflecting strong adaptability to heterogeneous modalities.
\textbf{d) Conclusion.} Overall, LargeMvC-Net shows leading performance across all large-scale scenarios, consistently outperforming shallow and deep competitors. 
Its unified and optimization-aware architecture ensures scalable clustering in both complete and incomplete multi-view environments.

\subsubsection{Intuitive Results} 

\begin{figure*}[t]
  \centering
  \includegraphics[width=0.9\linewidth]{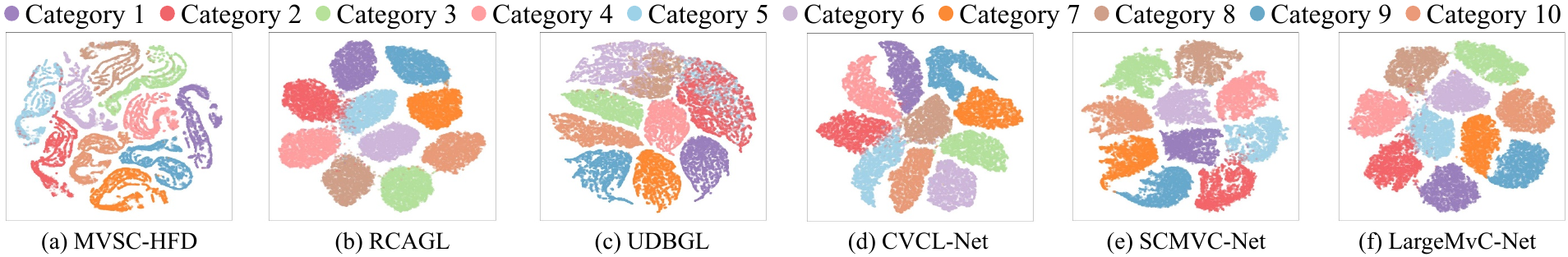}\\
  \caption{The t-SNE visualizations based on the clustering representations of Cifar10 dataset.}
  \label{tnserep}
\end{figure*}

Fig.~\ref{tnserep} (all in \textbf{Appendix} Fig. \ref{tnseCifar10}) presents the t-SNE visualizations of clustering results on Cifar10 dataset. 
Compared with other methods, LargeMvC-Net yields the most compact and well-separated clusters, clearly reflecting class boundaries and minimal overlaps. 
This aligns with our quantitative results, where LargeMvC-Net consistently achieves the highest clustering performance. 
The visual clarity further highlights the benefit of jointly optimizing representation and anchor indicators in an end-to-end unfolding framework.

\subsection{Component and Parameter Analysis}\label{sec:abla}

\begin{figure*}[t]
  \centering
  \includegraphics[width=0.9\linewidth]{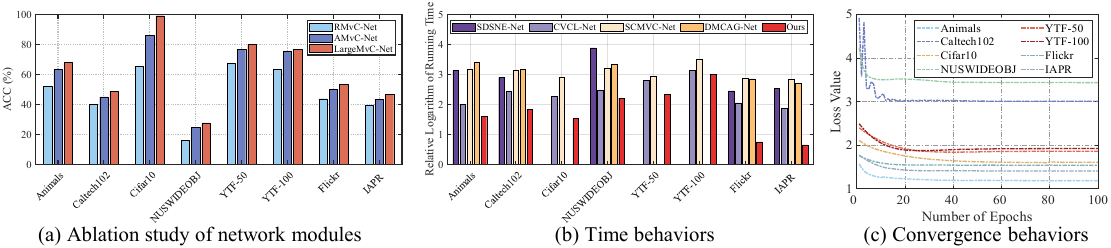}\\
  \caption{(a) Ablation study of network modules; (b) Time comparison with deep methods; (c) Convergence analysis.}
  \label{ablationtimeloss}
\end{figure*}

\subsubsection{Network Architecture Analysis} \label{networkablation}

Fig.~\ref{ablationtimeloss} (a) indicates the ablation results of different module combinations. 
RMvC-Net, which only includes RepresentModule, performs poorly due to the lack of structural alignment. 
Incorporating AnchorModule in AMvC-Net significantly improves clustering by enabling structural modeling and scalability. 
The full model, LargeMvC-Net, further introduces NoiseModule, leading to consistent performance gains across all datasets.
This highlights the benefit of jointly unfolding representation, noise, and anchor components.

\subsubsection{Time and Convergence Analysis} \label{Convergenceablation} 

First, Fig.~\ref{ablationtimeloss} (b) and Table \ref{DeepComplexity} in Subsection \ref{ConvergenceAnalysis} illustrate both empirical and theoretical efficiency. 
LargeMvC-Net achieves the lowest runtime and superior scalability by leveraging anchor-based decomposition and unfolding design.
This avoids expensive pairwise computations and scales linearly with sample size.
Moreover, as shown in Fig.~\ref{ablationtimeloss} (c), LargeMvC-Net demonstrates fast and stable convergence across all eight datasets, with most losses plateauing within 20 epochs. 
This reinforces the efficient optimization behavior of the model, as discussed in Subsection \ref{ComputationalComplexityAnalysis}.

\subsubsection{Parameter Analysis} \label{ParameterAnalysisRQ3}

\begin{figure}[t]
  \centering
  \includegraphics[width=\linewidth]{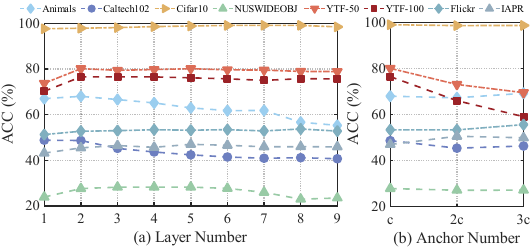}\\
  \caption{(a) Parameter sensitivity of layers and (b) anchors.}
  \label{LayerandAnchor}
\end{figure}

In Fig.~\ref{LayerandAnchor}, the performance generally increases as the number of layers (a) and the number of anchors (b) grow across all datasets, with the best performance achieved at 2 layers and $m = c$ anchors. 
This suggests that while deeper networks and more anchors improve clustering accuracy, further increases may lead to diminishing returns due to overfitting or computational complexity. 
The search in Subsection \ref{ImplementationDetails} ensures that the model effectively scales to large datasets while maintaining performance.


\section{Conclusion and Future Work}\label{sec:Conclusion}

In this paper, we bridged the gap in existing deep anchor-based multi-view clustering by unfolding the underlying optimization problem into a principled deep architecture, termed LargeMvC-Net.
The architecture's modular design includes representation learning, noise suppression, and anchor indicator estimation. 
Coupled with the unsupervised reconstruction loss, it ensures that the learned representations are expressive and well-aligned with each view’s structure.
Extensive experiments demonstrated that LargeMvC-Net outperformed state-of-the-art methods in both clustering quality and scalability across a variety of benchmarks.
Future work will focus on extending the model to handle more complex, large-scale multi-view data, such as incorporating dynamic anchor structures and exploring multi-task learning for multi-view clustering.

\begin{acks}
This work is in part supported by the National Natural Science Foundation of China under Grants U21A20472 and 62276065, and the Fujian Provincial Natural Science Foundation of China under Grant 2024J01510026.
\end{acks}
\bibliographystyle{ACM-Reference-Format}
\bibliography{ML}

\clearpage

\setcounter{section}{0}
\setcounter{table}{0}
\setcounter{figure}{0}
\renewcommand\thesection{\Alph{section}}

\section{Supplementary of Convergence Proof}\label{SupplementaryofOverallConvergence}

\begin{proof}\label{Proof1}
First, the network implicitly solves the following objective function as
\begin{equation}\label{allobj}
\begin{array}{ll}
&\mathcal{J}\left(\mathbf{H}^{t}, \{\mathbf{E}_{v}^{t}\}_{v=1}^{V}, \{\mathbf{P}_{v}^{t}\}_{v=1}^{V}\right) = \sum\limits_{{v=1}}^{V}\Big(\frac{1}{2}\Vert\mathbf{X}_{v}-\mathbf{H}\mathbf{P}_{v}-\mathbf{E}_{v}\Vert_{F}^{2}\\& +\alpha \Vert\mathbf{H}\Vert_{1} + \beta \Vert\mathbf{E}_{v}\Vert_{2, 1}\Big), \text { s.t. } \mathbf{H} \geq 0, \mathbf{P}_{v}\mathbf{P}_{v}^{\top}=\mathbf{I}.
\end{array}
\end{equation}
Function \eqref{allobj} is non-convex, so we design a three-step unfolding network based on alternating iteration optimization method to optimize it.
It can be decomposed into a set of sub-problems, each of which can be optimally solved.
The alternating optimization strategy ensures that the objective function decreases monotonically with each iteration and converges, as supported by the theoretical guarantees in \cite{Bezdek03Convergence}.
Taking the $t$-th training epoch solution $\mathbf{H}^{(t)}, \{\mathbf{E}_{v}^{(t)}\}_{v=1}^{V}, \{\mathbf{P}_{v}^{(t)}\}_{v=1}^{V}$ as an example, we can obtain the following process.

i) Given $\{\mathbf{E}_{v}^{(t)}\}_{v=1}^{V}, \{\mathbf{P}_{v}^{(t)}\}_{v=1}^{V}$, the optimal $\mathbf{H}^{(t+1)}$ can be analytically obtained. 
Suppose the optimal solution be $\mathbf{H}^{(t+1)}$, we have
\begin{equation}\label{allobj1}
\begin{array}{ll}
&\mathcal{J}\left(\mathbf{H}^{(t)}, \{\mathbf{E}_{v}^{(t)}\}_{v=1}^{V}, \{\mathbf{P}_{v}^{(t)}\}_{v=1}^{V}\right) \geq \\& \mathcal{J}\left(\mathbf{H}^{(t+1)}, \{\mathbf{E}_{v}^{(t)}\}_{v=1}^{V}, \{\mathbf{P}_{v}^{(t)}\}_{v=1}^{V}\right).
\end{array}
\end{equation}

ii) With $\mathbf{H}^{(t+1)}, \{\mathbf{P}_{v}^{(t)}\}_{v=1}^{V}$ fixed, the optimal solution $\{\mathbf{E}_{v}^{(t+1)}\}_{v=1}^{V}$ can be analytically derived.
This optimal solution update of $\{\mathbf{E}_{v}^{(t+1)}\}_{v=1}^{V}$ guarantees a non-increasing objective value as
\begin{equation}\label{allobj2}
\begin{array}{ll}
&\mathcal{J}\left(\mathbf{H}^{(t+1)}, \{\mathbf{E}_{v}^{(t)}\}_{v=1}^{V}, \{\mathbf{P}_{v}^{(t)}\}_{v=1}^{V}\right) \geq \\& \mathcal{J}\left(\mathbf{H}^{(t+1)}, \{\mathbf{E}_{v}^{(t+1)}\}_{v=1}^{V}, \{\mathbf{P}_{v}^{(t)}\}_{v=1}^{V}\right).
\end{array}
\end{equation}

iii) Keeping $\mathbf{H}^{(t+1)}, \{\mathbf{E}_{v}^{(t+1)}\}_{v=1}^{V}$ fixed, the optimal $\{\mathbf{P}_{v}^{(t+1)}\}_{v=1}^{V}$ update can be computed in closed form. 
This leads to a guaranteed decrease or maintenance of the optimal objective value of $\{\mathbf{P}_{v}^{(t+1)}\}_{v=1}^{V}$ as
\begin{equation}\label{allobj3}
\begin{array}{ll}
&\mathcal{J}\left(\mathbf{H}^{(t+1)}, \{\mathbf{E}_{v}^{(t+1)}\}_{v=1}^{V}, \{\mathbf{P}_{v}^{(t)}\}_{v=1}^{V}\right) \geq \\& \mathcal{J}\left(\mathbf{H}^{(t+1)}, \{\mathbf{E}_{v}^{(t+1)}\}_{v=1}^{V}, \{\mathbf{P}_{v}^{(t+1)}\}_{v=1}^{V}\right).
\end{array}
\end{equation}

By combining the three steps above, we obtain the following \textbf{Lemma \ref{Lemma1}}.
\begin{lemma}\label{Lemma1}
Each sub-problem update in the proposed optimization framework guarantees a non-increasing value of the objective function $\mathcal{J}(\cdot)$, \emph{i.e.,}
\begin{equation}\label{allobj4}
\begin{array}{ll}
&\mathcal{J}\left(\mathbf{H}^{(t)}, \{\mathbf{E}_{v}^{(t)}\}_{v=1}^{V}, \{\mathbf{P}_{v}^{(t)}\}_{v=1}^{V}\right) \geq \\& \mathcal{J}\left(\mathbf{H}^{(t+1)}, \{\mathbf{E}_{v}^{(t+1)}\}_{v=1}^{V}, \{\mathbf{P}_{v}^{(t+1)}\}_{v=1}^{V}\right).
\end{array}
\end{equation}
\end{lemma}
Then, we have \textbf{Theorem \ref{Theorem1}} and \textbf{Corollary \ref{Corollary1}} based on \textbf{Lemma \ref{Lemma1}} in the main paper.
Here, the proof is completed.
\end{proof}

\section{Discussion}\label{Discussionsup}

\subsection{Discussion on Training Losses}\label{DiscussiononTrainingLosses}
Unlike contrastive or pseudo-label based approaches, our reconstruction loss avoids reliance on handcrafted pretext tasks or noisy clustering signals.
Furthermore, this objective naturally enforces cross-view consistency: Since $\mathbf{H}$ is shared across all views and responsible for generating each $\mathbf{X}_{v}$ via $\mathbf{P}_{v}$, minimizing $\mathcal{L}_{R}$ encourages structurally coherent representations across heterogeneous feature spaces.
In practice, this loss is computed at the final unfolding layer of the network and propagated back through all modules, allowing RepresentModule, NoiseModule, and AnchorModule to be jointly trained via standard back-propagation. 
The resulting network is entirely unsupervised scheme, scalable to large-scale scenarios.
In addition to the reconstruction loss, our framework can be extended to incorporate auxiliary losses such as entropy regularization or neighborhood-preserving constraints to further refine clustering structure. 
However, we find that the core anchor-based reconstruction objective already provides strong performance and stability across benchmarks, as presented in Subsection~\ref{sec:exprq}.

\section{Supplementary of Experiments}\label{SupplementaryofExperiments}


\subsection{Details of Experimental Setups}\label{DetailsofExperimentalSetup}

\subsubsection{Datasets}\label{SupDatasetsandTestSettings}

Multi-view datasets include 
Animals\footnote{http://attributes.kyb.tuebingen.mpg.de/},
Caltech102\footnote{http://www.vision.caltech.edu/Image\_Datasets/Caltech101/}, Cifar10\footnote{http://www.cs.toronto.edu/kriz/cifar.html},
MNIST\footnote{http://yann.lecun.com/exdb/mnist/}, NUSWIDEOBJ\footnote{http://lms.comp.nus.edu.sg/research/NUS-WIDE.htm},
YouTubeFace, YTF-50 and YTF-100 are three versions of YouTubeFaces\footnote{https://www.cs.tau.ac.il/$\sim${wolf}/ytfaces/}, ESP-Game\footnote{https://www.kaggle.com/datasets/parhamsalar/espgame}, Flickr\footnote{https://press.liacs.nl/mirflickr/}, and IAPR\footnote{https://www.imageclef.org/photodata}.
These datasets corresponds to two types of scenarios: 1) Animals, Caltech102, Cifar10, MNIST, NUSWIDEOBJ, YTF-50 and YTF-100 datasets contain different manual and deep features; 2) Flickr and IAPR datasets include various vision and language features.
The statistics of these datasets are summarized below.
\textbf{1) Animals} is a deep multi-feature dataset that consists of 10,158 images from 50 animal classes with DECAF and VGG-19 features;
\textbf{2) Caltech102} is a popular object recognition dataset with 102 classes of images.
Six extracted features are available: Gabor, wavelet moments, CENTRIST, histogram of oriented gradients, GIST, and LBP features.
\textbf{3) Cifar10} consists of 50,000 tiny images that can be divided into ten mutually exclusive classes. 
We extract its features on DenseNet, ResNet101, ResNet50 networks.
\textbf{4) NUSWIDEOBJ} consists of 30,000 images distributed over 31 classes. 
We use five features provided by NUS, \emph{i.e.,} color histogram, color moments, color correlation, edge distribution, and wavelet texture features.
\textbf{5) YouTubeFaces (YTF)} is a large-scale database of face videos designed for studying the problem of unconstrained face recognition in videos, and we extend it as a series of multi-view facial datasets with 50 and 100 classes, including LBP, HOG, GIST, and Gabor features.
Moreover, YouTubeFace is also a dataset of YTF that has five views.
\textbf{6) ESP-Game} originates from an image annotation game played on a website, which contains 20,770 images, and each image is annotated by players with several descriptions. 
Here, we choose 11,032 images that are described with approximately five tags per image, and these images have a total of 7 classes.
\textbf{7) Flickr} provides 25,000 images with 1,386 text tags downloaded from the social photography site Flickr. 
We select 12,154 images across 7 categories for experiments.
\textbf{8) IAPR} is a public image dataset with 20,000 images in 6 classes, each with a short text description. 
After filtering out images with fewer than 4 tags, 7,855 images were randomly selected.
For ESP-Game, Flickr and IAPR datasets, image features are extracted using VGG-16, and text features using BERT, forming a multi-modal dataset.

\subsection{Supplementary of Ablation Models}\label{SupAblationmodel}

\begin{table*}[t]
\centering
\caption{The alternating iteration algorithm-based solution of objective \eqref{allproblem} guides the design of diverse ablation networks.}
\resizebox{\textwidth}{!}{
\begin{tabular}{c|c|l}
\toprule
\multicolumn{1}{c|}{Variant Models} & \multicolumn{1}{c|}{Concretized Anchor-based Clustering Optimization Problems}  & \multicolumn{1}{c}{Composition of Deep Unfolding Network Modules}\\
\midrule
RMvC-Net & $\mathop{\min}\limits_{\mathbf{H}}\sum\limits_{{v=1}}^{V}\Big(\frac{1}{2}\Vert\mathbf{X}_{v}-\mathbf{H}\mathbf{P}_{v}\Vert_{F}^{2}+\alpha \Vert\mathbf{H}\Vert_{1}\Big), \text { s.t. } \mathbf{H} \geq 0$ & 
\textbf{RepresentModule:} $\mathbf{H}^{(l+1)} \leftarrow \frac{1}{V}\sum\limits_{{v=1}}^{V}\left(\mathbfcal{S}_{\theta^{(l)}}\Big(\mathbf{H}^{(l)}\mathbf{R}+\mathbf{X}_{v}(\mathbf{P}_{v}^\top)^{(l)}\mathbf{U}\right)$ \\
\midrule
\multirow{2}{*}{AMvC-Net} & \multirow{2}{*}{$\mathop{\min}\limits_{\mathbf{H}, \mathbf{P}_{v}}\sum\limits_{{v=1}}^{V}\Big(\frac{1}{2}\Vert\mathbf{X}_{v}-\mathbf{H}\mathbf{P}_{v}\Vert_{F}^{2}+\alpha \Vert\mathbf{H}\Vert_{1}\Big), \text { s.t. } \mathbf{H} \geq 0, \mathbf{P}_{v}\mathbf{P}_{v}^{\top}=\mathbf{I}$} &
\textbf{RepresentModule:} $\mathbf{H}^{(l+1)} \leftarrow \frac{1}{V}\sum\limits_{{v=1}}^{V}\left(\mathbfcal{S}_{\theta^{(l)}}\Big(\mathbf{H}^{(l)}\mathbf{R}+\mathbf{X}_{v}(\mathbf{P}_{v}^\top)^{(l)}\mathbf{U}\right)$ \\
& & \textbf{AnchorModule:} $\mathbf{P}_{v}^{(l+1)}=\mathbf{B}_{v}^{(l+1)}(\mathbf{C}_{v}^{\top})^{(l+1)}$\\
\midrule
\multirow{3}{*}{LargeMvC-Net} & \multirow{3}*{$\mathop{\min}\limits_{\mathbf{H}, \mathbf{P}_{v}, \mathbf{E}_{v}}\sum\limits_{{v=1}}^{V}\Big(\frac{1}{2}\Vert\mathbf{X}_{v}-\mathbf{H}\mathbf{P}_{v}-\mathbf{E}_{v}\Vert_{F}^{2}+\alpha \Vert\mathbf{H}\Vert_{1} + \beta \Vert\mathbf{E}_{v}\Vert_{2, 1}\Big), \text { s.t. } \mathbf{H} \geq 0, \mathbf{P}_{v}\mathbf{P}_{v}^{\top}=\mathbf{I}$} &
\textbf{RepresentModule:} $\mathbf{H}^{(l+1)} \leftarrow \frac{1}{V}\sum\limits_{{v=1}}^{V}\left(\mathbfcal{S}_{\theta^{(l)}}\Big(\mathbf{H}^{(l)}\mathbf{R}+(\mathbf{X}_{v}-\mathbf{E}_{v}^{(l)})(\mathbf{P}_{v}^\top)^{(l)}\mathbf{U}\right)$ \\
& & \textbf{NoiseModule:} $\mathbf{E}_{v}^{(l+1)}\leftarrow\mathbfcal{D}_{\rho_{v}^{(l)}}\left(\mathbf{X}_{v}- \mathbf{H}^{(l+1)}\mathbf{P}_{v}^{(l)}\right)$ \\
& & \textbf{AnchorModule:} $\mathbf{P}_{v}^{(l+1)}=\mathbf{B}_{v}^{(l+1)}(\mathbf{C}_{v}^{\top})^{(l+1)}$\\
\bottomrule
\end{tabular}}
\label{NetworkExamples}
\end{table*}

\begin{figure*}[t]
  \centering
  \includegraphics[width=\linewidth]{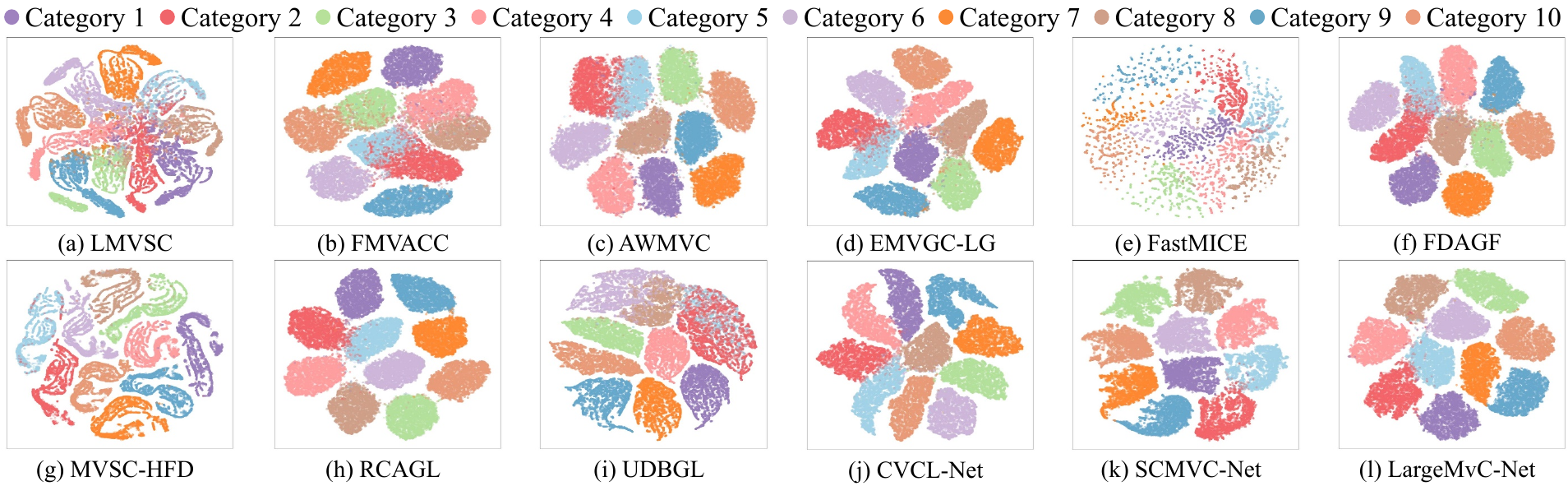}\\
  \caption{The t-SNE visualizations based on the clustering representations of Cifar10 dataset.}
  \label{tnseCifar10}
\end{figure*}

The introduction of the ablation models are shown in Table \ref{NetworkExamples} and below.
\begin{itemize}
\item \textbf{Ablation Model 1}: RepresentMvC-Net (RMvC-Net) is an ablated variant that retains only RepresentModule from the full optimization-inspired framework.
\item \textbf{Ablation Model 2}: AnchorMvC-Net (AMvC-Net) introduces anchor structures into RMvC-Net, serving as a scalable variant for large-scale clustering environments.
\item \textbf{Final Model}: Based on the previous variants, LargeMvC-Net further introduces noise variables into the optimization, completing the full model design.
\end{itemize}

\subsection{Supplementary of Experimental Results}\label{ExperimentalSupplementaryofExperimentalResults}

Fig. \ref{tnseCifar10} supplements the t-SNE visualization results of all representations on Cifar10 dataset.

\end{document}